\documentclass[10pt,twocolumn]{article} 
\usepackage{simpleConference}
\usepackage{times}
\usepackage{graphicx}
\usepackage{amssymb}

\usepackage{authblk}
\usepackage{epsfig}
\usepackage{amsmath}
\usepackage{array,multirow,graphicx}
\usepackage{makecell}
\usepackage{pifont}
\usepackage{enumitem}
\usepackage{placeins}
\usepackage{balance}
\usepackage{lipsum}
\usepackage{subfiles}

\usepackage[pagebackref=true,breaklinks=true,letterpaper=true,colorlinks,bookmarks=false]{hyperref}
\usepackage[margin=10pt,font=normalsize,labelfont=md]{caption}
\newcommand{\Tau}{\mathrm{T}}
\newcommand{\Kappa}{\mathrm{K}}
\newcommand{\Beta}{\mathrm{B}}

\begin{document}

\title{DEFT: Detection Embeddings for Tracking}

\author[2, 3]{Mohamed Chaabane}
\author[1]{Peter Zhang}
\author[3]{J. Ross Beveridge}
\author[2]{Stephen O'Hara}
\affil[1]{Uber ATG, Louisville, CO}
\affil[2]{Aurora, Louisville, CO}
\affil[3]{Colorado State University, Fort Collins, CO}
\date{\vspace{-5ex}}

\maketitle
\thispagestyle{empty}

\begin{abstract}
Most modern multiple object tracking (MOT) systems follow the tracking-by-detection paradigm, consisting of a detector followed by a method for associating detections into tracks. There is a long history in tracking of combining motion and appearance features to provide robustness to occlusions and other challenges, but typically this comes with the trade-off of a more complex and slower implementation. Recent successes on popular 2D tracking benchmarks indicate that top-scores can be achieved using a state-of-the-art detector and relatively simple associations relying on single-frame spatial offsets -- notably outperforming contemporary methods that leverage learned appearance features to help re-identify lost tracks. In this paper, we propose an efficient joint detection and tracking model named DEFT, or ``Detection Embeddings for Tracking." Our approach relies on an appearance-based object matching network jointly-learned with an underlying object detection network. An LSTM is also added to capture motion constraints. DEFT has comparable accuracy and speed to the top methods on 2D online tracking leaderboards while having significant advantages in robustness when applied to more challenging tracking data. DEFT raises the bar on the nuScenes monocular 3D tracking challenge, more than doubling the performance of the previous top method. Code is publicly available.\footnote{ \url{https://github.com/MedChaabane/DEFT}}

\end{abstract}

\section{Introduction}

Visual Multi-Object Tracking (MOT) has made significant progress in recent years, motivated in part from high-profile mobile robotics and autonomous driving applications. Continued improvements in the accuracy and efficiency of Convolutional Neural Network (CNN) based object detectors has driven the dominance of the ``tracking by detection" paradigm \cite{bewley2016simple,tang2017multiple,xu2019spatial,ciaparrone2020deep,bergmann2019tracking}. Recent work has shown that simple tracking mechanisms added to state of the art detectors \cite{Zhou2020TrackingOA} can outperform more complex trackers reliant upon older detection architectures.


The tracking by detection approach is characterized by two main steps: 1) detection of objects in a single video frame, and 2) association to link objects detected in the current frame with those from previous frames. There are many ways to associate detections across frames, but those featuring learned associations are interesting because they have the promise of addressing edge-cases where modeling and heuristics-based approaches fail. Even with learned associations, the two-stage  approach can lead to sub-optimal results in terms of accuracy and efficiency. A recent trend of jointly learning detection and tracking tasks in a single neural network has led to increases in performance on tracking benchmarks and related applications. However, existing end-to-end methods that combine appearance and motion cues can be complex and slow (see \S\ref{section:related}, Related Work).



We posit that a learned object matching module can be added to most contemporary CNN-based object detectors to yield a high performing multi-object tracker, and further, that by jointly training the detection and tracking (association) modules, both modules adapt to each other and together perform better. Using the same backbone for object detection and inter-frame association increases efficiency and accuracy when compared to methods that use detection as a black-box feeding input to the association logic. 

In this paper we present an approach in which embeddings for each object are extracted from the multi-scale backbone of a detector network and used as appearance features in an object-to-track association sub-network. We name our approach ``Detection Embeddings for Tracking" (DEFT). We show that DEFT can be applied effectively to several popular object detection backbones. Due to the gains of feature sharing in the network design, our approach using both appearance and motion cues for tracking has speed comparable to leading methods that use simpler association strategies. Because DEFT keeps a memory of appearance embeddings over time, it is more robust to occlusions and large inter-frame displacements than the top alternatives. This robustness allows DEFT to dramatically outperform competing methods on the challenging nuScenes 3D monocular vision tracking benchmark.



\section{Related Work}
\label{section:related}

\textbf{Tracking-by-detection.} Most state-of-the-art trackers follow the approach of tracking-by-detection which heavily relies on the performance of the detector. Trackers in this approach often use detectors as black box modules and focus only on associating detections across frames. In the pre-deep learning era, trackers often used Kalman filters \cite{bewley2016simple,chen2018real}, Intersection-over-Union (IoU) \cite{bewley2016simple,bochinski2017high} or flow fields \cite{rodriguez2009tracking,izadinia20122} for association. These methods are simple and fast but they fail very easily in challenging scenarios. Recently, with the success of deep learning, many models have used appearance features to associate objects \cite{ciaparrone2020deep}. DeepSORT \cite{wojke2017simple} for example, takes provided detections and associates them using an offline trained deep re-identification (ReID) model and Kalman filter motion model. SiameseCNN \cite{leal2016learning} uses a Siamese network to directly learn the similarity between a pair of detections. The Deep Affinity Network of Sun et al. \cite{sun2019deep} employs a Siamese network that takes 2 video frames as input, extracts multi-scale appearance embeddings and outputs similarity scores between all pairs of detections. Mahmoudi et al. \cite{mahmoudi2019multi} use extracted visual features alongside dynamic and position features for object association. In \cite{bae2017confidence}, the authors combined the output of the CNN with shape and motion models. The main disadvantage of these models is that they use time-consuming feature extractors and they treat detection and association separately, leading to suboptimal accuracy and speed. In DEFT, association is jointly learned with detection in a unified network. Feature extraction for matching reuses the detection backbone, thus object associations are computed with only a small additional latency beyond detection. 

\textbf{Joint detection and tracking.} The recent success of multi-task learning in deep neural networks \cite{chaabane2020looking,trabelsi2021pose,zhang2017survey} has led to models that jointly learn detection and tracking tasks. Tracktor \cite{bergmann2019tracking} adapts the Faster RCNN detector \cite{renNIPS15fasterrcnn} to estimate the location of a bounding box in the new frame from the previous frame. Tracktor's drawback is that it works well only on high frame-rate videos where inter-frame motion is low. In \cite{feichtenhofer2017detect}, authors extend R-FCN \cite{dai2016r} detector to compute correlation maps between high level feature maps of consecutive frames to estimate inter-frame offsets between bounding boxes. Similarly, CenterTrack \cite{Zhou2020TrackingOA} extends the CenterNet detector \cite{zhou2019objects} to estimate inter-frame offsets of bounding boxes. CenterTrack is a state-of-the-art method, but only associates objects in consecutive frames. Our method is more robust for longer occlusions and large inter-frame displacements, thus improving tracking under more challenging conditions, which we show in \S\ref{section:nuScenesAnalysis}. Xu et al. \cite{xu2020train} presents an end-to-end MOT training framework, using a differentiable approximation of MOT metrics in the loss functions. They show improvements for existing deep MOT methods when extended with their training framework. Chaabane et al. \cite{chaabane2021end} propose an approach to jointly optimize detection and tracking, with a focus on static object tracking and geolocalization. Their model makes strong use of learned pose estimation features, and is less suitable as a general tracker for dynamic objects. JDE \cite{wang2019towards} extends YOLOv3 \cite{redmon2018yolov3} with a reID branch to extract object embeddings for association. The feature extractor and detection branches share features and are jointly learned. Similarly, FairMOT \cite{zhang2020fairmot} improves on JDE and makes use of the CenterNet detector to improve tracking accuracy. 

DEFT is similar in approach to JDE and FairMOT. These methods jointly learn detection and feature matching in a single network as the basis for visual multiple object tracking. DEFT provides additional evidence that jointly learning detection and matching features can provide a simple and effective (SOTA) tracking solution. DEFT overcomes some of the limitations of competing approaches such as CenterTrack when applied to challenging examples by providing a longer track memory over which similarity scores are aggregated and by applying a simple LSTM-based motion model to filter physically implausible matches. 


\section{DEFT Network}
\begin{figure*}[t]
\begin{center}
  \includegraphics[width=\linewidth]{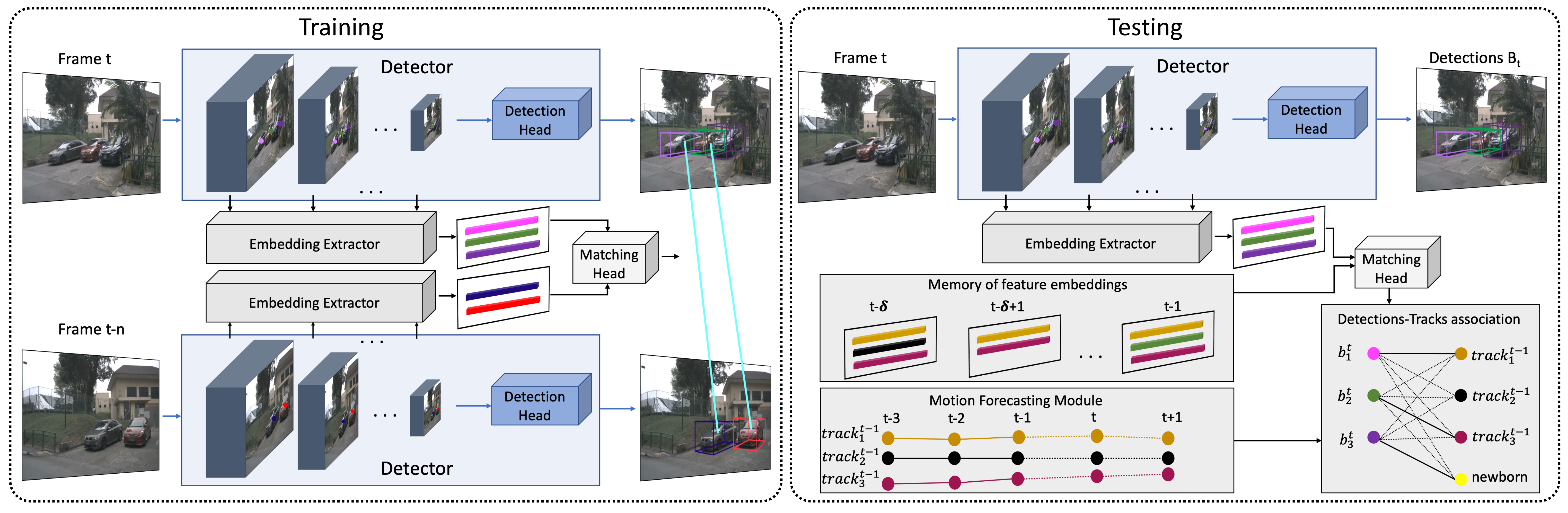}
\end{center}
  \caption{DEFT Training and Inference. DEFT adds an embedding extractor and a matching head to an object detection backbone to jointly train appearance features for both detection and association tasks. During inference, the matching head matches the current detections to the embeddings remembered for all active tracks and uses a motion forecasting module to eliminate implausible trajectories.}
\label{diagram}
\end{figure*}

Given the tracking-by-detection paradigm, we propose to exploit the representational power of the intermediate feature maps of the object detector (``detector backbone") to extract embeddings to be used in an object matching sub-network that associates objects across frames. We jointly train the detector and the object matching network. During training, errors in object association propagate back through the detection backbone such that the appearance features are optimized for both detection and matching. DEFT also employs a low-dimensional LSTM module to provide geometric constraints to the object matching network. DEFT implemented with a CenterNet \cite{zhou2019objects} backbone achieves state-of-the-art performance on several tracking benchmarks, while being faster than most similarly-scoring alternatives. The speedup is in part due to the fact that in DEFT, object association is a small additional module within the detection network, thus adding only a few milliseconds of latency beyond detection.

During inference (See Figure \ref{diagram}), the embedding extractor head uses features maps and bounding boxes from the detector as input, and extracts appearance embeddings for each detected object. The matching head uses the embeddings to compute a similarity between the objects in the current frame and those remembered from previous frames (current tracks). A motion forecasting module (LSTM) prevents matches that lead to physically implausible trajectories. The Hungarian Algorithm is used to make the final online association of objects to tracks. Details for each module are provided below, followed by the training procedure.

\subsection{Object Embeddings}

The object embeddings used in the matching network are extracted from the detection backbone, as described below. We label this the ``embedding extractor" in Figure \ref{diagram} and subsequent text. The embedding extractor constructs representative embeddings from the intermediate feature maps of the detector backbone to help associate (or, ``re-identify") objects during tracking. We use feature maps at different layers to extract appearance from multiple receptive fields (RFs), which provides additional robustness over single-RF embeddings. DEFT takes a video frame $t$ as input and, via the detection head, outputs a set of bounding boxes $\Beta_t=\{b^t_1,b^t_2,...,b^t_{N_t}\}$. For convenience, we use $N_t = \vert\Beta_t\vert$ to represent the number of bounding boxes in frame $t$.


For each detected object, we extract feature embeddings from the estimated 2D object's center location. For 3D bounding boxes, we use projection of 3D center location into the image space as its estimated 2D center location. If the center of the $i^{\textrm{th}}$ object is at position $(x,y)$ in the input frame of size $W \times H$, then for a feature map of size $W_m \times H_m \times C_m$, we extract the $C_m$-dimensional vector at position $(\frac{y}{H}H_m,\frac{x}{W}W_m )$ as the feature vector $f^m_i$ for the $i^{\textrm{th}}$ object in feature map $m$. We concatenate features from $M$ feature maps to construct the resulting $e$-dimensional feature embedding $f_i = f^1_i \cdot f^2_i \dots f^M_i $  for the $i^{\textrm{th}}$ object. 

The dimension of the feature vector $f^m_i$ affects its contribution to the resulting feature embedding. To change the contribution of some feature maps and to control the dimension of the embedding, we add a single convolution layer to some feature maps to increase/decrease the dimension from $C_m$ to $C^{\prime}_{m}$ before extracting the feature vectors. In practice, this serves to increase the dimension of features contributed by earlier maps while reducing those from later maps. More details on this can be found in the supplementary material.

\subsection{Matching Head}

The matching head follows the Deep Affinity Network \cite{sun2019deep}, using the object embeddings to estimate the similarity scores between all pairs of detections across the two frames. With $N_{max}$ maximum number of allowed objects in each frame, we construct the tensor $E_{t,t-n} \in \mathbb{R}^{N_{max} \times N_{max} \times 2e} $ such that the feature embedding of each object in frame $t$ is concatenated along the depth dimension with each feature embedding from objects in frame $t-n$ and vice versa. To construct fixed size tensor $E_{t,t-n}$ we pad the rest of the tensor with zeros. $ E_{t,t-n} $ is fed to the matching head which is composed of a few (4-6) layers of $1 \times 1$ convolutions. The output of the matching head is affinity matrix $A_{t,t-n} \in \mathbb{R}^{N_{max} \times N_{max}} $. 

Since we learn similarity between embeddings, there is no guarantee that the resulting scores are symmetric when matching backwards or forwards across frames, i.e., when matching from ($t \rightarrow t-n$) versus ($t-n \rightarrow t$). As a result, we compute the affinities in both directions using a separate affinity matrix for each, denoted with superscripts ``bwd" and ``fwd" in the following.

To allow for objects that should not be associated between the frames (objects new to the scene, or departed), we add a column to $A_{t,t-n}$ filled with constant value $c$. We apply softmax to each row to obtain matrix $\hat{A}^{bwd}$, representing the final affinity including non-matched scores. The choice of $c$ is not overly sensitive -- the network will learn to assign affinities greater than $c$ for true matches.

Each $\hat{A}^{bwd}[i,j]$, represents the estimated probability of associating $b^t_i$ to $b^{t-n}_j$. $\hat{A}^{bwd}[i, N_{max}+1]$ represents the estimated probability that $b^t_i$ is an object not present in frame $t-n$. Similarly, we construct the forward affinity matrix, using the transpose matrix  $A^T_{t,t-n}$ to which we add a column filled with constant value $c$ and then we apply softmax to each row to obtain matrix $\hat{A}^{fwd}$. During the inference, the similarity score between $b^t_i$ and $b^{t-n}_j$ is given as the average of $\hat{A}^{bwd}[i,j]$ and $\hat{A}^{fwd}[j,i]$.


\subsection{Online Data Association} 
\label{association}

In DEFT, tracks remember the object embeddings from each observation comprising the track in the last $\delta$ frames. The association between a new detection and the existing tracks requires computing the similarity of the new object to the set of observations from each track in memory. To allow for occlusions and missed detections, track memory is maintained for a few seconds so that inactive tracks can be revived if a new observation is strongly associated to the previous observations. Tracks with no observations after $N_{age}$ frames are discarded from memory.

We define a track $\Tau$ as the set of associated detections from frame $t-\delta$ to $t-1$, noting that tracks may not have a detection for each frame. The size of the track is given as $\vert\Tau\vert$, representing the number of bounding boxes and associated embeddings it comprises. We define the distance between the $i^{\textrm{th}}$ new detection $b^t_i$ and $\Tau_j$ as:

\begin{equation}
d({b^t_i},\Tau_j) = \frac{1}{\vert\Tau_j\vert} \sum^{}_{b_k^{t-n} \in \Tau_j} \frac{\hat{A}^{fwd}_{t,t-n}[k,i] + \hat{A}^{bwd}_{t,t-n}[i,k] }{2}
\label{eqn:detection-to-track}
\end{equation}

The detections-to-tracks association problem is formulated as a bipartite matching problem so that exclusive correspondence is guaranteed. Let $\Kappa = \{\Tau_j\}$ be the set of current tracks. We construct the detections-to-tracks similarity matrix $D \in \mathbb{R}^{\vert\Kappa\vert \times (N_t + \vert\Kappa\vert)}$ by appending the all-pairs detections-to-tracks distance (Eq. \eqref{eqn:detection-to-track}) of size $\vert\Kappa\vert \times N_t$ with a matrix $X$ of size $\vert\Kappa\vert \times \vert\Kappa\vert$ used to represent when a track is associated with no detections in the current frame. The entries along the diagonal of $X$ are computed as the average non-match score of the detections in the track, the off-diagonal entries are set to $-\infty$. Specifically, $D$ is constructed as follows:

{
\scriptsize
\begin{flalign}
D & = [S \vert X] \\
S[j,i] & =  d(b^t_i,\Tau_j) \\
X[j,k] & = 
\begin{cases}
    \scriptsize
    \frac{1}{\vert\Tau_j\vert} \sum^{}_{b_k^{t-n} \in \Tau_j}\hat{A}^{fwd}_{t,t-n}[k,N_{max}+1] & j = k\\
    -\infty, & j \neq k \\
\end{cases}
\end{flalign}
} 

%

Finally, we solve the bipartite matching problem defined by $D$ with the Hungarian algorithm \cite{kuhn1955hungarian}. We include only likely associations if the affinity is larger than a specified threshold $\gamma_1$. Unmatched detections will start newborn tracks. Tracks that have not been associated for more than a predefined maximum age $N_{age}$ are considered to have left the scene and are deleted from the track set.

\subsection{Motion Forecasting}
\label{motion-forcasting}
When learning to associate detections across frames using the appearance features from a detection backbone, there is some chance that two objects look similar enough in the embedding space to cause confusion. It is common practice to add additional geometric or temporal constraints to help resolve such ambiguities. This often takes the form of a Kalman Filter (e.g. \cite{wojke2017simple,chen2018recurrent}) or an LSTM module (e.g. \cite{sadeghian2017tracking}). DEFT uses an LSTM as our motion forecasting module. This module predicts future locations of each track in the next $\Delta T_{\textrm{pred}}$ frames given its information in $\Delta T_{\textrm{past}}$ past frames. Motion forecasting is used to constrain the associations between frames to those that are physically plausible. The motion forecasting module sets affinity scores in Eq. \eqref{eqn:detection-to-track} to $-\infty$ for detections that are too distant from the track's predicted location. Further details on the LSTM implementation are in the supplementary material. In \S\ref{section-ablation}, we provide an ablation study to show the impact of our LSTM motion forecasting module, and we also show that it modestly outperforms a Kalman Filter in our application.


\subsection{Training}
\label{section:training}

During training, a pair of frames, $n$ frames apart, is input to DEFT as shown in Figure \ref{diagram}.  Following \cite{chaabane2020end}, the image pairs are separated by a random number of frames $1 \leq n \leq n_{gap}$ to encourage the network to learn to be robust to temporary occlusions or missed detections.


For each training pair, we create two ground truth matching matrices $M^{fwd}$ and $M^{bwd}$, representing the forward and backward associations, respectively. The ground truth matching matrices consist of entries $[i,j] \in \{0, 1\}$, and have dimensions ${N_{max} \times (N_{max}+1)}$ to allow for unassociated objects. A value of $1$ in the matrix encodes an association, and is also used for unassociated objects. Everywhere else, the value is $0$.





To train DEFT for matching estimation, we use the loss function $\mathcal{L}_{\textrm{match}}$ defined as the average of the two losses $\mathcal{L}^{fwd}_{\textrm{match}}$ and $\mathcal{L}^{bwd}_{\textrm{match}}$, where $\mathcal{L}^{bwd}_{\textrm{match}}$ is the error of matching bounding boxes in frame $t$ to those in frame $t-n$ and $\mathcal{L}^{fwd}_{\textrm{match}}$ is the error of matching bounding boxes in frame $t-n$ to those in frame $t$. The expression of the matching loss is given as follows, where ``*" represents ``fwd" or ``bwd" as appropriate:

\begin{eqnarray} \label{eqAffinityLoss}
\mathcal{L}^{*}_{\textrm{match}} & = & \sum^{N_{max}}_{i=1} \sum^{N_{max}+1}_{j=1}  M^*[i,j] \textrm{log}(\hat{A}^*[i,j]), \\
\mathcal{L}_{\textrm{match}} & = &  \frac{\mathcal{L}^{fwd}_{\textrm{match}} + \mathcal{L}^{bwd}_{\textrm{match}}}{2 (N_t + N_{t-n})},  
\end{eqnarray}

Training optimizes the joint affinity and detection losses as defined in Eq. \eqref{eqjoint}. For a better optimization of our proposed dual-task network, we use the strategy proposed in \cite{kendall2018multi} for automatic loss balancing the two tasks. 
 
\begin{equation} \label{eqjoint}
\mathcal{L}_{\textrm{joint}}=   \frac{1}{e^{\lambda_1}}  (\frac{ \mathcal{L}^t_{\textrm{detect}} +  \mathcal{L}^{t-n}_{\textrm{detect}}}{2}) +   \frac{1}{e^{\lambda_2}} \mathcal{L}_{\textrm{match}} + \lambda_1 + \lambda_2
\end{equation}

where $ \mathcal{L}^t_{\textrm{detect}}$ is the detection loss for frame $t$ and $\lambda_1$ and $\lambda_2$ are the balancing weights to the two tasks. Note that the balancing weights are modeled as learnable parameters. 

\section{Experiments}
\label{section:experiments}

\subsection{Datasets and Metrics}

We evaluated the tracking performance of DEFT on a set of popular benchmarks: MOT Challenge (MOT16/MOT17) \cite{milan2016mot16}, KITTI tracking \cite{geiger2012we}, and the nuScenes Vision Tracking benchmark \cite{caesar2020nuscenes}. The MOT Challenge and KITTI benchmarks are used to assess 2D visual tracking, while nuScenes is used for monocular 3D visual tracking.

\textbf{MOT16/MOT17.} The MOT16 and MOT17 tracking challenges are part of the multi-object tracking benchmark MOT Challenge \cite{milan2016mot16}. They are composed of indoor and outdoor pedestrian tracking sequences.
The videos have frame-rates between 14 and 30 FPS. Both challenges contain the same seven training sequences and seven test sequences. We test our tracker using public detections provided by the benchmark protocol, following \cite{bergmann2019tracking}, as well as with private detections output by DEFT. With public detections, we use the jointly-trained DEFT model, but employ the provided bounding boxes for embedding extraction -- all else is the same. With private detections, we use the bounding boxes output from our model.

MOT Challenge benchmarks employ the following metrics: MOTA - multi-object tracking accuracy, MOTP - multi-object tracking precision, IDF1 - identity F1 Score, MT - mostly tracked, ML - mostly lost, FP - false positives, FN - false negatives, IDS - identity switches. We refer the reader to \cite{bernardin2008evaluating, ristani2016performance} for detailed definitions. 


\textbf{KITTI.} The KITTI tracking benchmark is composed of 21 training sequences and 29 test sequences that were collected using cameras mounted on top of a moving vehicle. The videos are recorded at 10 FPS. We evaluate performance using the ``Car" class because it is the only class with enough examples to allow effective training without external data sources.\footnote{This shouldn't be interpreted as a lack of generality, as our results on MOT17 and nuScenes show.} Public detections are not provided with KITTI. KITTI uses the same tracking metrics as the MOT Challenge benchmarks.

\textbf{nuScenes.} nuScenes is a large-scale data set for autonomous driving. nuScenes is composed of 1000 sequences, with 700, 150, 150 sequences for train, validation, and testing, respectively. Sequences were collected in Boston and Singapore, in both day and night, and with different weather conditions. Each sequence length is roughly 20 seconds with camera frequency of 12 FPS. Each sequence contains data from six cameras forming full 360$^{\circ}$ field of view but box annotations are provided only for key frames (2 FPS). Given the ground truth format, only key frames are used for training and evaluation. The effectively low frame-rate of 2FPS makes this data set challenging for visual tracking as the inter-frame motion of objects can be large. We evaluate DEFT on the 7 annotated classes: Car, Truck, Trailer, Pedestrian, Bicycle, Motorcycle, Bus. nuScenes uses tracking metrics aggregated over the curve of operating thresholds \cite{weng2019baseline}. They are as follows: AMOTA - average MOTA, AMOTP - average MOTP, MOTAR - recall-normalized MOTA score.


\subsection{Implementation Details}

DEFT is implemented using PyTorch with an open source license. In all experiments, we used an Ubuntu server with a TITAN X GPU with 12 GB of memory. All hyper-parameters were chosen based on the best MOTA score for 3-fold cross validation for 2D tracking and best AMOTA score on the validation set for 3D tracking. Our implementation runs at approximately 12.5Hz on all data sets. 

\begin{table}
\footnotesize
\begin{center}
\begin{tabular}{lcccc}
\hline
DEFT Variant & MOTA$\uparrow$ & FP $\downarrow$ & FN $\downarrow$ & IDS $\downarrow$\\
\hline
DEFT + YOLO v3 \cite{redmon2018yolov3} &85.6 &6.5\%  &6.8\% &1.1\%\\
DEFT + FPN \cite{lin2017feature} &87.0 &6.0\%  &6.3\% &0.7\%\\
DEFT + Faster R-CNN \cite{renNIPS15fasterrcnn} &87.7 &5.8\%  &\textbf{5.8}\% &0.7\%\\
DEFT + CenterNet \cite{zhou2019objects} &\textbf{88.1} &\textbf{5.7}\%  &5.9\% &\textbf{0.3}\%\\
\hline
\end{tabular}
\end{center}
\caption{3-fold cross-validation results of implementing DEFT with different object detector networks on KITTI.}
\label{DEFT_variants}
\end{table}

\textbf{2D Tracking.} 
DEFT was trained and evaluated with four object detectors including CenterNet \cite{zhou2019objects}, YOLO v3 \cite{redmon2018yolov3}, FPN \cite{lin2017feature} and Faster R-CNN \cite{renNIPS15fasterrcnn}. All object detectors were pre-trained on the COCO dataset \cite{lin2014microsoft}. We augmented the data with random cropping, flipping, scaling and photometric distortions. Table~\ref{DEFT_variants} shows the relative performance from 3-fold cross validation on KITTI. We also performed this evaluation using MOT17, which yielded the same ranking, not shown here for brevity. Since the DEFT+CenterNet variant was the strongest performing, we use it as the detection backbone for the remaining results in this paper.

While DEFT achieves best performance with CenterNet, these results demonstrate that DEFT can achieve good tracking performance using different detector backbones. Interestingly, Faster R-CNN and CenterNet have similar detection performance on KITTI, however the association performance is better with CenterNet (fewer ID switches). This might be due the nature of CenterNet being an anchor-free detector and matches well to DEFT's design of using feature embeddings from the object center locations -- allowing the features at center locations to benefit from the supervised signals of both tasks.


When training DEFT with CenterNet for the various experiments in this paper, we used the following procedure. We used CenterNet with the modified DLA-34 \cite{zhou2019objects} backbone. We train DEFT for 80 epochs with a starting learning rate of $e^{-4}$ using the Adam \cite{kingma2014adam} optimizer, and batch size 8. The learning rate is reduced by 5 at epochs 30, 60, and 70. We extract feature embeddings from the 13 feature maps of the modified DLA-34 backbone. 

\begin{table}
\footnotesize
\begin{center}
\begin{tabular}{lccccccccc}
\hline
 Method &MOTA$\uparrow$ &MOTP $\uparrow$ &IDF1 $\uparrow$ &IDS $\downarrow$ \\
 \hline \hline 
 Tracktor17 \cite{bergmann2019tracking}& 53.5 &78.0 &52.3 &2072\\
 DeepMOT-Tracktor \cite{xu2020train}& 53.7 &77.2 &53.8 &1947\\
 Tracktor v2 \cite{bergmann2019tracking}& 56.3 &\textbf{78.8} &55.1 &1987\\
 GSM Tracktor \cite{liugsm}& 56.4 &77.9 &57.8 &\textbf{1485}\\
 CenterTrack \cite{Zhou2020TrackingOA} & \textbf{61.5} & -- & 59.6  & 2583 \\
 Ours (Public) & 60.4 & 78.1 & \textbf{59.7} &2581 \\
\hline \hline
CenterTrack (Private) & 67.8 & -- & 64.7  & 3039 \\
Ours (Private) & 66.6 & 78.83 & 65.42 & 2823\\
\hline 
\end{tabular}
\end{center}
\caption{\textbf{MOT17}. Comparison uses public detections (provided bounding boxes) and private detections. Compared methods are from the MOT leaderboard, except CenterTrack where the results are from their paper.}

\label{table_mot17}
\end{table}

\begin{table}[t]
\footnotesize
\begin{center}
\begin{tabular}{lccccccc}
\hline
Method &MOTA$\uparrow$ &MOTP $\uparrow$ &MT $\uparrow$& ML $\downarrow$ &IDS $\downarrow$\\
\hline\hline
SMAT \cite{gonzalez2020smat}& 84.27 & \textbf{86.09} &63.08 &5.38 &341 \\
mono3DT \cite{hu2019joint}& 84.52 &85.64  &73.38 &2.77 &377\\
mmMOT \cite{zhang2019robust}& 84.77 &85.21 &73.23 &2.77 &284 \\
MASS \cite{karunasekera2019multiple}& 85.04 &85.53 &74.31 &2.77 & 301 \\
TuSimple \cite{choi2015near}& 86.62 &83.97 &72.46 &6.77 &293 \\
CenterTrack \cite{Zhou2020TrackingOA}&  \textbf{89.44} &85.84 &82.31 &2.31 &\textbf{116}\\
\hline 
Ours& 88.95 &84.55 &  \textbf{84.77} &  \textbf{1.85} &343 \\
\hline
\end{tabular}
\end{center}
\caption{\textbf{KITTI} car tracking. We compare to published online entries on the leaderboard.}
\label{kitti_table}
\end{table}


\textbf{3D Tracking.} 
The nuScenes evaluation was performed on the full 360$^{\circ}$ panorama and not with each camera separately. Following \cite{Zhou2020TrackingOA}, we fused outputs from all cameras naively without any additional post-processing for handling duplicate detections between cameras or for associating objects across cameras. This ensures a fair comparison between all monocular 3D trackers. We performed additional analysis on the validation set allowing us to observe how performance varies when controlling for certain variables, including the amount of occlusion in the track and a measure of inter-frame displacement. 


\textbf{Parameter Settings}
See the supplemental for the parameter settings we used to support each benchmark.

\subsection{Comparative Evaluation}


\begin{table}[ht]
\small
\begin{center}
\begin{tabular}{lccc}
\hline
Method & AMOTA & MOTAR & MOTA \\
\hline\hline
Mapillary \cite{simonelli2019disentangling}+ AB3D \cite{weng2019baseline} & 1.8 & 9.1 & 2.0  \\
CenterTrack \cite{Zhou2020TrackingOA} & 4.6 & 23.1 & 4.3 \\
\hline \hline 
Ours& \textbf{17.7} & \textbf{48.4} & \textbf{15.6}\\
\hline
\end{tabular}
\end{center}
\caption{\textbf{nuScenes Vision Tracking}. We compare to published monocular 3D tracking entries on the leaderboard.}
\label{nuscenes_table}
\end{table}

We compare against other online tracking methods using the protocol appropriate for each benchmark. We follow the common practice of comparing against published/peer-reviewed methods listed on the leaderboard. Tracking results for MOT, KITTI, and nuScenes benchmarks are computed by host test servers with hidden labels on the test set.

Both CenterTrack and DEFT use a CenterNet detection backbone. We see in Table~\ref{table_mot17} and Table~\ref{kitti_table} that both outperform the other published online trackers on the leaderboards, with a slight edge in performance to CenterTrack. This shows the power of the CenterNet detection backbone and provides support that jointly optimized detection and tracking methods can outperform those that have detection as a distinct step. DEFT achieves the best trajectory coverage on KITTI, providing evidence that DEFT maintains longer tracks better, possibly due to remembering embeddings for several observations in a track.

We observe a big performance advantage with DEFT on nuScenes (Table \ref{nuscenes_table}), which we attribute in large part due to differences in how well DEFT tolerates long occlusions and large inter-frame displacements of tracked objects. We explore this hypothesis in \S\ref{section:nuScenesAnalysis}. Compared with CenterTrack, DEFT achieves gain of 13.1 percentage points in AMOTA, 25.3 in MOTAR, and 11.3 in MOTA. As others have noted \cite{caesar2020nuscenes,Yang_2020_CVPR,zhu2019class}, nuScenes is significantly more difficult and closer to real-world scenarios than KITTI.

\begin{figure}[b]
\begin{center}
  \includegraphics[width=1.0\linewidth]{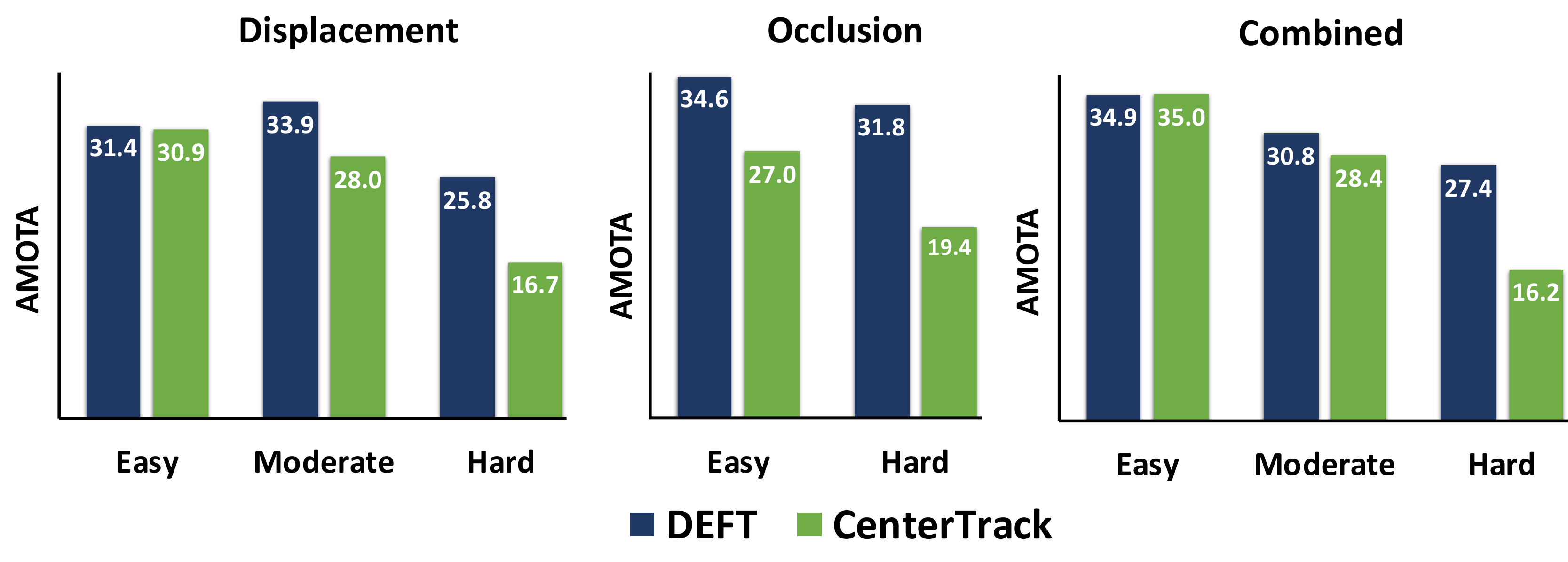}
\end{center}
  \caption{DEFT compared with CenterTrack on nuScenes validation front camera videos, according to difficulty factors of occlusion and inter-frame displacement.}
\label{fig:nuscenes_analysis}
\end{figure}

\begin{figure*}[t]
\begin{center}
  \includegraphics[width=0.79\linewidth]{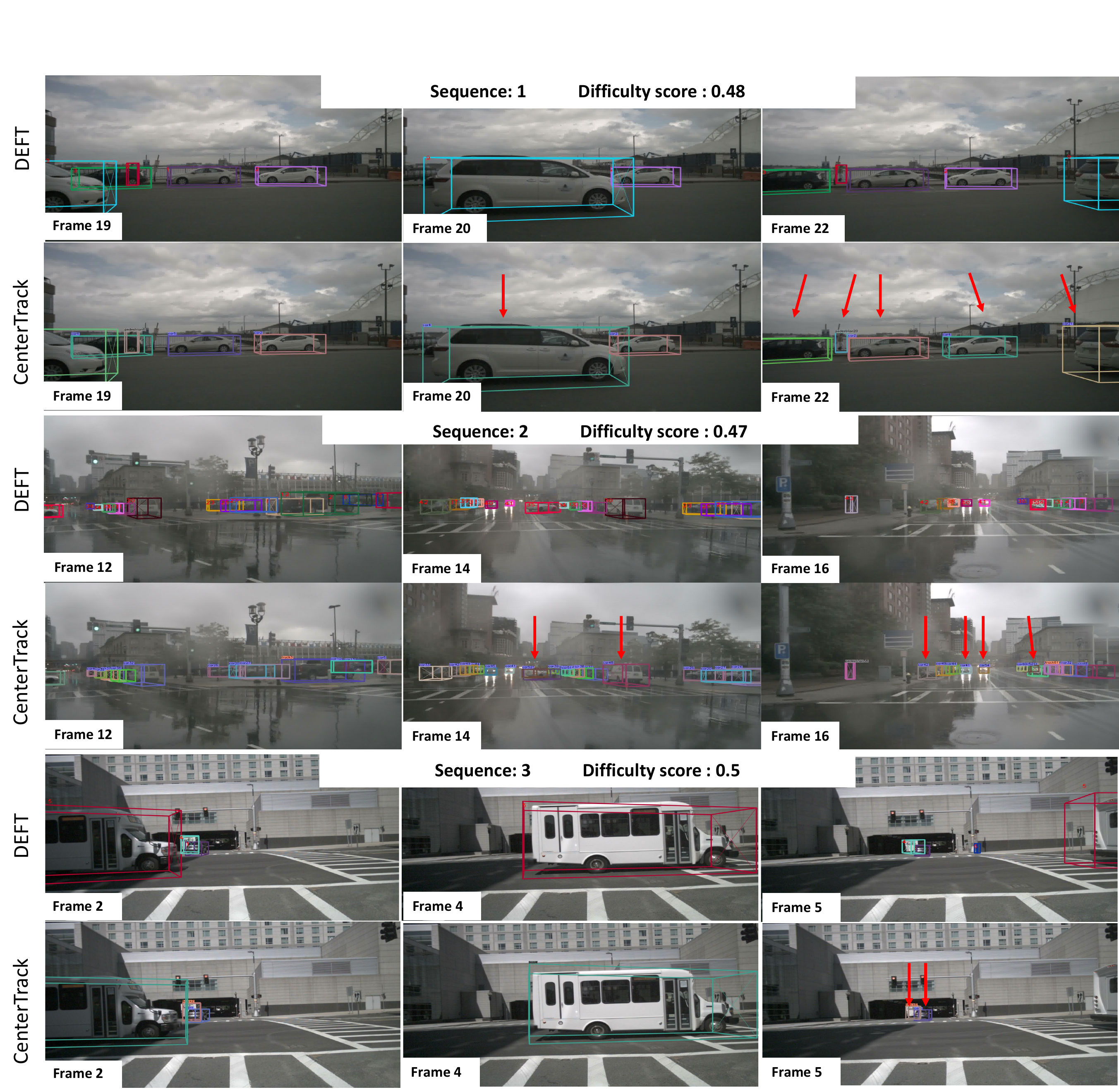}
\end{center}
  \caption{Qualitative results comparison between DEFT and CenterTrack \cite{Zhou2020TrackingOA} on nuScenes. Each pair of rows shows the results comparison for one sequence. The color of the boxes represents the identity of the tracks. Red arrows point at tracking errors (identity switches). Notice that DEFT is more robust to occlusions and large inter-frame displacements.}
\label{qualitative}
\end{figure*}

\subsection{Performance Analysis}
\label{section:nuScenesAnalysis}

As previously stated, DEFT and CenterTrack perform similarly on 2D tracking benchmarks of MOT17 and KITTI, but DEFT outscores CenterTrack and all other methods on the nuScenes vision tracking leaderboard by a sizable margin. In this section we investigate what factors may explain the performance difference.

Our intuition is that the major performance gains on KITTI and MOT benchmarks are driven by improved detectors. For MOT and KITTI, the tracking/association logic can be weak and still produce top numbers. Others in the community seem to agree. The authors of Tracktor promote ``tracking without bells and whistles" \cite{bergmann2019tracking}, while the creators of CenterTrack state that it ``...trades the ability to reconnect long-range tracks for simplicity, speed, and high accuracy in the local regime...this trade-off is well worth it." \cite{Zhou2020TrackingOA} The first row in Table~\ref{ablation_study} of our ablation study (see \S\ref{section-ablation}) provides additional support for this point of view. We observe that a simple baseline, using nothing other than a motion model and IOU associations, when applied to CenterNet detections, yields a MOTA score of 86.7 on KITTI (validation) and 63.5 on MOT17 (validation). While one cannot compare validation scores directly to test results, this is suggestive that many of the top leaderboard methods are only marginally better than a naive baseline coupled to a SOTA detector.

\begin{table*}[ht!]
\scriptsize
\begin{center}
\begin{tabular}{lccccccccccccccc}
\hline
&&&&\multicolumn{4}{c}{MOT17}&&\multicolumn{4}{c}{KITTI}&&\multicolumn{2}{c}{NuScenes}\\
\thead{\scriptsize{Feature}\\ \scriptsize{Embeddings}}   & \thead{\scriptsize{Motion}\\ \scriptsize{Model}}  & \thead{\scriptsize{2D/3D IOU}\\  \scriptsize{}} &&MOTA$\uparrow$ &MT $\uparrow$ &ML $\downarrow$ &IDS $\downarrow$&&MOTA$\uparrow$ &MT $\uparrow$ &ML $\downarrow$ &IDS $\downarrow$ && AMOTA$\uparrow$&MOTA$\uparrow$ \\ \cline{1-3} \cline{5-8} \cline{10-13} \cline{15-16} 
None & LSTM & \checkmark  && 63.5 & 19.6\%  & 38.1\%  &2.8\% & & 86.7 &51.3\% &27.9\% & 1.7\%&& 4.2 &5.0 \\ 
\hline
Single-Scale   &  &   &&64.0 &28.8\%  &28.5\%&2.3\% &&87.4 &81.46\%  &3.15\% &1.0\% && 17.1&15.0 \\ 
\hline
Multi-Scale   &  &   &&64.4 &29.8\%  &27.5\%  &1.9\% &&87.7 &82.55\%  &2.71\% &0.7\% && 18.4&16.2 \\ 
\hline
Multi-Scale   & Kalman &  &&65.2 &30.0\%  &27.3\%  &1.1\% &&87.8 &82.60\%  &2.66\% &0.6\% && 18.9&16.4 \\
\hline
Multi-Scale   & LSTM &  &&65.2 &30.0\%  &27.3\%  &1.1\% &&88.0 &82.81\%  &2.57\% &0.4\% && 20.0&17.2 \\ 
\hline
Multi-Scale   & LSTM &  \checkmark &&\textbf{65.4} &\textbf{30.3}\%  &\textbf{27.0\%}  &\textbf{0.9}\% &&\textbf{88.1} &\textbf{83.05}\%  &\textbf{2.36}\% &\textbf{0.3}\% && \textbf{20.9} &\textbf{17.8} \\ 
\hline
\end{tabular}
\end{center}
\caption{Ablation study of DEFT on MOT17, KITTI and nuScenes datasets. Results are obtained with 3-fold cross-validation on the training sets for MOT17 and KITTI, for nuScenes the results are on the validation set.}
\label{ablation_study}
\end{table*}

However, is tracking without bells and whistles sufficient for harder tasks? We divided the nuScenes validation data into partitions based on two factors, an occlusion score and a displacement score. The occlusion score for a track is the number of frames for which the object was temporarily occluded. We sum this over all tracks to score a video. The displacement score for a track is the mean of the 2D center displacements in consecutive frames. We use the mean of the top ten tracks as the video's score. Scores are linearly rescaled to the range $[0,1]$. The distribution of scores led us to divide the occlusion factor into easy and hard categories (below/above the median respectively), and the displacement factor into approximately equal-sized easy/moderate/hard partitions.\footnote{See supplemental for additional details.} We also look at a combined difficulty factor, which is simply the maximum of the two normalized difficulty scores.

Figure~\ref{fig:nuscenes_analysis} shows that the difference in performance between DEFT and CenterTrack is marginal on easier videos with respect to the displacement factor or the overall difficulty factor. CenterTrack outperforms DEFT on the easiest videos by less than one percentage point, which is consistent with the difference in performance we observe on KITTI and MOT benchmarks. However, when looking at moderate and hard videos, we observe that DEFT provides superior performance. For occlusions, CenterTrack drops 7.6 percentage points from easy to hard, whereas DEFT drops only 2.8, indicating that CenterTrack is more sensitive to occlusions than DEFT. Figure~\ref{qualitative} provides example frames from videos that feature occlusions and large displacements, where DEFT is more robust.

\subsection{Ablation Studies}
\label{section-ablation}

In Table~\ref{ablation_study}, we present the results of an ablation study on MOT17, KITTI and nuScenes benchmarks investigating the importance of various aspects of DEFT. The first row of the table shows baseline performance for tracking when using CenterNet detections followed by a simple non-learned motion+IOU association step. We observe that the baseline tracker performs reasonably well in MOT17 and KITTI validation data, but fails to perform well on the more challenging nuScenes test. 

The next two rows compare the benefit of using feature embeddings extracted from different resolutions in the detector backbone versus using only the features from the final feature map. We controlled for dimensionality, so that the embeddings are the same size for both the single-scale and multi-scale variants. Using multi-scale embeddings leads to modest improvements across the board. 

The table also shows the effect of the motion forecasting module -- having one improves results modestly on MOT17 and KITTI, with a stronger effect on nuScenes. An alternative to the learned LSTM motion model would be to use a standard Kalman filter. We found the performance of the LSTM over the Kalman filter to be more pronounced in nuScenes. Finally, there are a few situations in which having an IOU-based association layer as a second-stage to object matching can provide a small performance gain. This is shown in the final row of the table.

On MOT and KITTI, the naive baseline performs well-enough that additional gains from DEFT result in a cumulative benefit of only a couple percentage points across most metrics, with the exception of MT and ML (mostly tracked, mostly lost) scores. There we see a big jump in performance from the baseline to multi-scale DEFT. This suggests that learned appearance-based detection-to-track associations help maintain longer tracks. On nuScenes, the value of DEFT is obvious. The gains in AMOTA from baseline to multi-scale DEFT is over 14 percentage points, with an additional 1.5 points gained by adding the motion forecasting LSTM module. This confirms that the learned matching-based association step is critical to overall performance, and that the motion model is a helpful addition, but relatively minor in terms of the overall performance.

\section{Conclusion}
Most state of the art trackers on popular public benchmarks follow the tracking-by-detection paradigm, with substantial boosts in performance attributable in large part to improved object detectors. This has allowed top-scoring algorithms to use limited matching strategies while achieving high tracking performance and efficiency. The concept of tracking in the ``local regime," that is, constraining the association logic to relatively short temporal and spatial extents, has been shown to be effective on at least two popular 2D tracking benchmarks (MOT, KITTI). However, not all tracking challenges are ones where the assumption holds true that the local regime dominates performance. In self-driving car applications, objects tracked in side-mounted cameras experience large inter-frame displacements, and occlusions lasting a few seconds are not uncommon. Additionally, there are use-cases for tracking with lower frame-rate videos in bandwidth constrained domains.


We have shown that detection embeddings used with learned similarity scores provide an effective signal for tracking objects, and are more robust to occlusions and high inter-frame displacements. On KITTI and MOT tracking benchmarks, DEFT is comparable in both accuracy and speed to leading methods. On the more challenging nuScenes visual tracking benchmark,
tracking performance more than doubles compared to the previous state of the art, CenterTrack (3.8x on AMOTA, 2.1x on MOTAR). Further, DEFT and CenterTrack perform near parity when occlusions and inter-frame displacements are low. 
However, when either factor becomes more challenging, DEFT performs better. Importantly, these are not corner cases -- the moderate and hard difficulty samples represent the majority in nuScenes, not the minority. DEFT's significant improvement in these cases is of considerable practical significance. 

Ongoing work includes applying DEFT to LiDAR based detectors and fused LiDAR+Vision detectors, and measuring performance on additional data sets and applications.
\newpage

\twocolumn[{%
 \centering
 \huge Supplementary Material: \\ DEFT: Detection Embeddings for Tracking\\ [1.5em]
 \Large Mohamed Chaabane, Peter Zhang, J. Ross Beveridge, Stephen O'Hara \\ [3.0em]
}]

\setcounter{equation}{0}
\setcounter{figure}{0}
\setcounter{table}{0}
\setcounter{page}{1}
\setcounter{section}{0}
\makeatletter

\section{Motion Forecasting Module}

The motion forecasting module predicts future locations of each track in the next $\Delta T_{\textrm{pred}}$ frames given its information in $\Delta T_{\textrm{past}}$ past frames. It uses features from past bounding boxes coordinates for each track. These features are presented differently for 2D and 3D tracking.  \\

\textbf{2D Tracking.} For 2D tracking, features for each detection at time $t$ are represented by 8-dimensional vector $(x_t,y_t,w_t,h_t,v^x_t,v^y_t,\dfrac{\Delta_{w_t}}{\Delta_t},\dfrac{\Delta_{h_t}}{\Delta_t})$ containing the 2D center location, height,width, velocity in the $x$ and $y$ directions and change in width and height divided by time difference between consecutive detections . \\

\textbf{3D Tracking.} For 3D tracking, features for each detection at time $t$ are represented by 11-dimensional vector $(x_t,y_t,z_t,w_t,h_t,l_t,r_t,v^x_t,v^y_t,v^z_t,v^r_t)$ containing 3D the center location, height,width,length,orientation about the z-axis,  velocity in the $x$, $y$ and $z$ directions and the rotational velocity $v^r_t$.

\section{Parameter Settings}
\textbf{MOT16/MOT17.} For MOT16 and MOT17, frames are resized to $960 \times 544$. The embedding extractor head outputs a feature embedding of $e=416$ features. The hyper-parameters $N_{max}$, $n_{gap}$, $\delta$, $\Delta T_{\textrm{IoU}}$, $\gamma_1$, $\gamma_2$, $N_{age}$, $c$ were set to 100, 60, 50, 5, 0.1, 0.4, 50, 10 respectively. For the motion forecasting module, $\Delta T_{\textrm{past}}$, $\Delta T_{\textrm{pred}} $ were set to 15 and 10.\\ 

\textbf{KITTI.} For KITTI, the embedding extractor head outputs a feature embedding of $e=672$ features. The hyper-parameters $N_{max}$, $n_{gap}$, $\delta$, $\Delta T_{\textrm{IoU}}$, $\gamma_1$, $\gamma_2$, $N_{age}$, $c$ were set to 100, 30, 25, 3, 0.1, 0.6, 30, 10 respectively. For the motion forecasting module, $\Delta T_{\textrm{past}}$, $\Delta T_{\textrm{pred}} $ were set to 10 and 5.\\

\textbf{nuScenes}
For 3D monocular tracking in nuScenes, DEFT was trained and evaluated with CenterNet as 3D object detector backbone. The embedding extractor head outputs a feature embedding of $e=704$ features. The frames are resized to $800 \times 448$. The hyper-parameters $N_{max}$, $n_{gap}$, $\delta$, $\Delta T_{\textrm{IoU}}$, $\gamma_1$, $\gamma_2$, $N_{age}$, $c$ were set to 100, 6, 5, 1, 0.1, 0.2, 6, 10 respectively. For motion forecasting module, $\Delta T_{\textrm{past}}$, $\Delta T_{\textrm{pred}} $ were set to 10 and 4.


\section{Detectors Implementation Details}

DEFT was trained and evaluated with four object detectors including CenterNet \cite{zhou2019objects}, YOLO v3 \cite{redmon2018yolov3}, FPN \cite{lin2017feature} and Faster R-CNN \cite{renNIPS15fasterrcnn}. All object detectors were pre-trained on the COCO dataset \cite{lin2014microsoft}. We followed same implementation details and hyper-parameters settings from their official public codes. With all detectors, the embedding extractor head outputs a feature embedding of $e=416$ and $e=672$  for MOT and KITTI respectively. For nuScenes, We only train and test with CenterNet and $e$ is set to 704.\\

\textbf{CenterNet.} We used CenterNet with the modified DLA-34 \cite{yu2018deep} backbone. We extract feature embeddings from all 13 feature maps of the modified DLA-34 backbone. More details of the backbone can be found in \cite{zhou2019objects}. \\

\begin{table*}[t!]
\begin{center}
\begin{tabular}{lccccccccccc}
\hline
&&\multicolumn{5}{c}{KITTI}&&\multicolumn{3}{c}{nuScenes}\\
Method &&MOTA &IDs  &AP$_E$  &AP$_M$ &AP$_H$&&AMOTA &MOTA &  mAP  \\ \cline{1-1} \cline{3-7} \cline{9-11} 
DEFT (separate training) &&87.9&0.5\% &92.8 &83.6&\textbf{74.3} &&20.1  &17.1 &24.5\\
DEFT (joint training)  &&\textbf{88.1}&\textbf{0.3}\% &92.8 &\textbf{83.8}&74.2 &&\textbf{20.9}  &\textbf{17.8} &24.5\\
\hline
\end{tabular}
\end{center}
\caption{Tracking and detection results of implementing DEFT with two training strategies (jointly vs separately optimized) on KITTI and nuScenes. Results are obtained with 3-fold cross-validation for KITTI where detection is evaluated with 2D bounding box AP for three different difficulty levels: easy (AP$_E$),  moderate (AP$_M$) and  hard (AP$_H$). Results are obtained on the validation set for nuScenes where detection is evaluated with mean Average Precision (mAP) over all 7 classes. }
\label{training_joint_vs_separate}
\end{table*}

\begin{table*}[t!]
\begin{center}
\begin{tabular}{lccccccccc}
\hline
Method &MOTA$\uparrow$ &MOTP $\uparrow$ &IDF1 $\uparrow$ &MT $\uparrow$& ML $\downarrow$ &IDS $\downarrow$ & Hz$\uparrow$ \\
\hline\hline
 Tracktor17 \cite{bergmann2019tracking} & 54.4 &78.2 &52.5 &19.0 &36.9 &682 & 1.5\\
 DeepMOT-Tracktor \cite{xu2020train} & 54.8 &77.5 &53.4 &19.1 &37.0 &645 &4.9 \\
  Tracktor v2 \cite{bergmann2019tracking} & 56.2 & \textbf{79.2} & 54.9 &20.7 &35.8 &617 &1.6\\
  GSM Tracktor \cite{liugsm} & 57.0 &78.1 &58.2 &22.0 &34.5 &\textbf{475} &8.7\\
  Ours (Public) & \textbf{61.7} &78.3 & \textbf{60.2} & \textbf{27.0} & \textbf{31.8} &768  &\textbf{12.3}\\
  \hline \hline
JDE \cite{wang2019towards} (Private) & 64.4 & - &55.8 &35.4 & 20.0 &1544 &22.2\\
Ours (Private) & 68.03 &78.71 &66.39 &33.06 &22.92 &925 &12.3\\
\hline
\end{tabular}
\end{center}
\caption{\textbf{MOT16}. We present the results of our approach with using public (provided) and private detections. JDE is not present on the public leaderboard, results are from their paper.}
\label{table_mot16}
\end{table*}

\textbf{YOLO v3.} We used YOLO v3 with Darknet-53 backbone. Darrknet-53 is composed of 53 convolutional layers. We extract feature embeddings from 12 features maps which are the output of layers 4, 10, 14, 19, 23, 27, 31, 35, 39, 44, 48 and 52 from Darknet-53.  \\

\textbf{FPN and Faster R-CNN.} For FPN and Faster R-CNN, we used ResNet101 \cite{he2016deep} backbone. We extract feature embeddings from 11 features maps which are the output of layers 7, 10, 16, 22, 34, 46, 58, 70, 82, 91 and 100 from ResNet101.

\section{Joint vs Separate Training}
To show the benefit of joint training, we compare DEFT with joint and separate training strategies. We can see from Table~\ref{training_joint_vs_separate} that when jointly training detection and tracking tasks, tracking performance is improved on both KITTI and nuScenes datasets without hurting the detection performance.

\section{MOT16 Results}

The MOT16 tracking challenge is part of the multi-object tracking benchmark MOT Challenge \cite{milan2016mot16}. MOT16 contains the same seven training sequences and seven test sequences as MOT17, as described in the main paper. MOT16 results are generally similar to MOT17. Including them here allows a comparison to additional methods.

Table~\ref{table_mot16} shows the accumulated results over all sequences of MOT16. One contemporary method that shares some conceptual similarity to DEFT is JDE, which has peer-reviewed results on MOT16, but is not listed on the public leaderboard. We include the results from their paper for comparison.



\section{Difficulty Splits on nuScenes}

\begin{figure}[h]
\begin{center}
  \includegraphics[width=\linewidth]{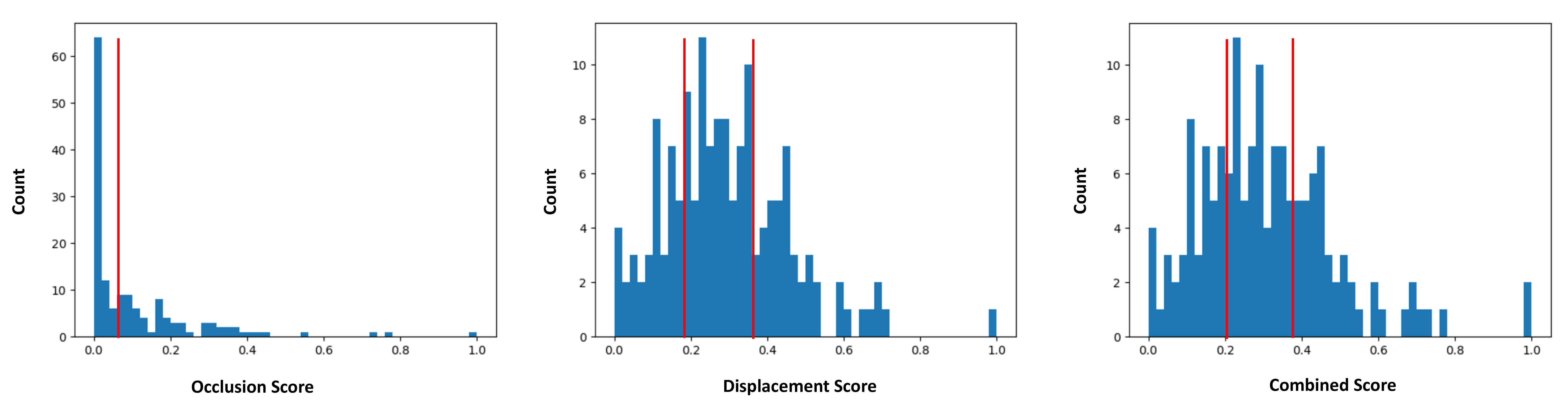}
\end{center}
  \caption{Occlusion, Displacement, and Combined scores distribution. Red lines represent thresholds used for the split.}
\label{difficulty_splits}
\end{figure}

We divided the nuScenes validation data into partitions based on three factors: an occlusion score, a displacement score and a combined score. Figure \ref{difficulty_splits} shows the scores distribution. We divided the occlusion factor into 76 easy and 74 hard videos (below/above the median respectively). We divided the displacement factor based on two thresholds of half standard deviation from the median score to obtain 44 easy, 55 moderate and 51 hard videos. Similarly, we divide the combined score into 43 easy,60 moderate and 47 hard videos.




\section{Additional 3D Monocular Tracking results}
Here we provide a per-class breakdown of performance when evaluating tracking performance on the nuScenes validation data, front-camera only imagery. The Trailer class is particularly difficult when using private (vision-based) detections. When using the provided lidar-based detections, tracking performance becomes more consistent with the other classes. This points out that some classes may not have enough training samples to train a robust visual detector, lacking the lidar signal.

\begin{table}[h]
\small
\begin{center}
\begin{tabular}{lccc}
\hline
 & CenterTrack \cite{Zhou2020TrackingOA}& Ours & Ours (Public Det) \\
\hline
 Bicycle& 16.7& 20.6 & 27.3\\
 Bus& 39.7& 46.4 & 66.1\\
 Car& 49.6& 54.6 & 71.9\\
 Motorcycle& 23.8 & 28.4 & 50.5\\
 Pedestrian& 31.4& 38.9 & 69.5\\
 Trailer& 0.0 & 0.3 & 47.2\\
 Truck& 15.1& 20.4 & 48.0\\
 \hline \hline
 Overall& 25.2 & 30.6 & 54.3\\
\hline
\end{tabular}
\end{center}
\caption{nuScenes 3D monocular Tracking results on the validation set. We present the results of our approach with private detections (those from our network) and public detections (from the lidar-based MEGVII \cite{zhu2019class} ) }
\label{DEFT_variants2}
\end{table}

\clearpage

{\small
\bibliographystyle{ieee_fullname}
\balance
\bibliography{refs}
}

\end{document}